\documentclass[10pt,twocolumn,letterpaper]{article}

\usepackage{iccv}
\usepackage{times}
\usepackage{epsfig}
\usepackage{graphicx}
\usepackage{amsmath}
\usepackage{amssymb}

\usepackage{pifont}
\newcommand{\cmark}{\ding{51}}%
\newcommand{\xmark}{\ding{55}}%

\usepackage{url}
\usepackage{tabularx}
\usepackage{enumitem}
\setlist{nosep}
\usepackage{soul}

\usepackage{balance}
\usepackage{subfigure}
\usepackage[multiple]{footmisc}
\usepackage{authblk}
\usepackage[table]{xcolor}
\usepackage{colortbl}
\definecolor{lightgray}{gray}{0.9}

\newcommand{\highlight}[1]{\textcolor{black}{#1}}

\usepackage{float}
\usepackage{stfloats}

\usepackage{multirow}

\makeatletter
\renewcommand\AB@affilsepx{, \protect\Affilfont}
\makeatother

\usepackage[pagebackref=true,breaklinks=true,letterpaper=true,colorlinks,bookmarks=false]{hyperref}

\iccvfinalcopy 

\def\iccvPaperID{2285} 

\ificcvfinal\fi

\begin{document}

\title{OpenForensics: Large-Scale Challenging Dataset\\For Multi-Face Forgery Detection And Segmentation In-The-Wild}

\author[1]{Trung-Nghia Le}
\author[2]{Huy H. Nguyen}
\author[2]{Junichi Yamagishi}
\author[2,3]{Isao Echizen}

\affil[1]{\small{National Institute of Informatics}} 
\affil[2]{\small{The Graduate University for Advanced Studies (SOKENDAI)}} 
\affil[3]{\small{University of Tokyo}}

\makeatletter
\g@addto@macro\@maketitle{
\vspace{-13mm}
\centering \textbf{\url{https://sites.google.com/view/ltnghia/research/openforensics/}}
\vspace{-3mm}
  \begin{figure}[H]
  \setlength{\linewidth}{\textwidth}
  \setlength{\hsize}{\textwidth}
  \centering
  \includegraphics[width=1\linewidth]{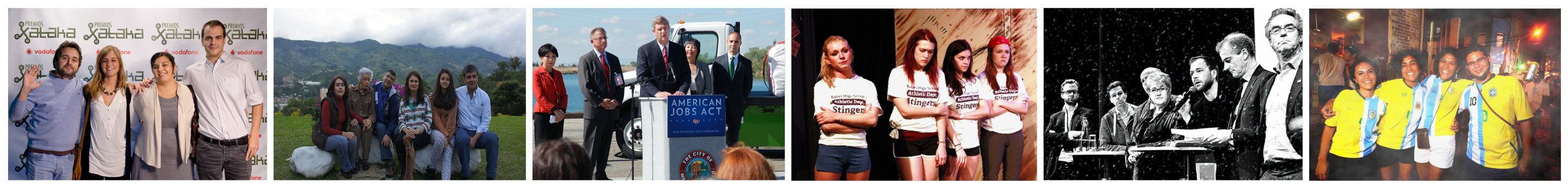}
	\caption{Examples from our OpenForensics dataset (best viewed online in color with zoom-in). Can you spot the forged faces and identify the manipulated areas in these images? The answers are in the supplementary material.}
	\label{fig:examples}
  \end{figure}
}
\makeatother

\maketitle

\begin{abstract}

The proliferation of deepfake media is raising concerns among the public and relevant authorities. It has become essential to develop countermeasures against forged faces in social media. This paper presents a comprehensive study on two new countermeasure tasks: multi-face forgery detection and segmentation in-the-wild. Localizing forged faces among multiple human faces in \highlight{unrestricted} natural scenes is far more challenging than the traditional deepfake recognition task. To promote these new tasks, we have created the first large-scale dataset posing a high level of challenges that is designed with face-wise rich annotations explicitly for face forgery detection and segmentation, namely OpenForensics. With its rich annotations, our OpenForensics dataset has great potentials for research in both deepfake prevention and general human face detection. We have also developed a suite of benchmarks for these tasks by conducting an extensive evaluation of state-of-the-art instance detection and segmentation methods on our newly constructed dataset in various scenarios. 
\end{abstract}
\vspace{-5mm}
\section{Introduction}
\label{sec:introduction}

Continuing advances in deep learning have led to impressive improvements in deepfake methods (\ie, deep learning-based face forgery), which can change the target person's identity~\cite{faceswap, deepfakes, faceswap-gan, Lingzhi-CVPR2020}. Emerging techniques such as autoencoder (AE) models and generative adversarial networks (GANs) enable transferring one person’s face to another person while retaining the original facial expression and head pose~\cite{Thies-CVPR2016, Thies-TG2019, Nirkin-ICCV2019, Zhixin-ECCV2018}. The realistic appearance synthesized with deepfake methods is drawing much attention in the fields of computer vision and graphics because of the potential application of such methods in a wide range of areas~\cite{faceapp, reface, Kim-TG2018, Egor-ICCV2019, deepfake_edu}. Moreover, falsified AI-synthesized images/videos have raised serious concerns about individual harassment and criminal deception~\cite{deepfake_porn, deepnude, deepfake_scam}. To address threats posed by spoofing and impersonation attacks, it is essential to develop countermeasures against face forgeries in digital media. 

\begin{figure}[t!]
    \vspace{-1mm}
	\centering
	\includegraphics[width=1\linewidth]{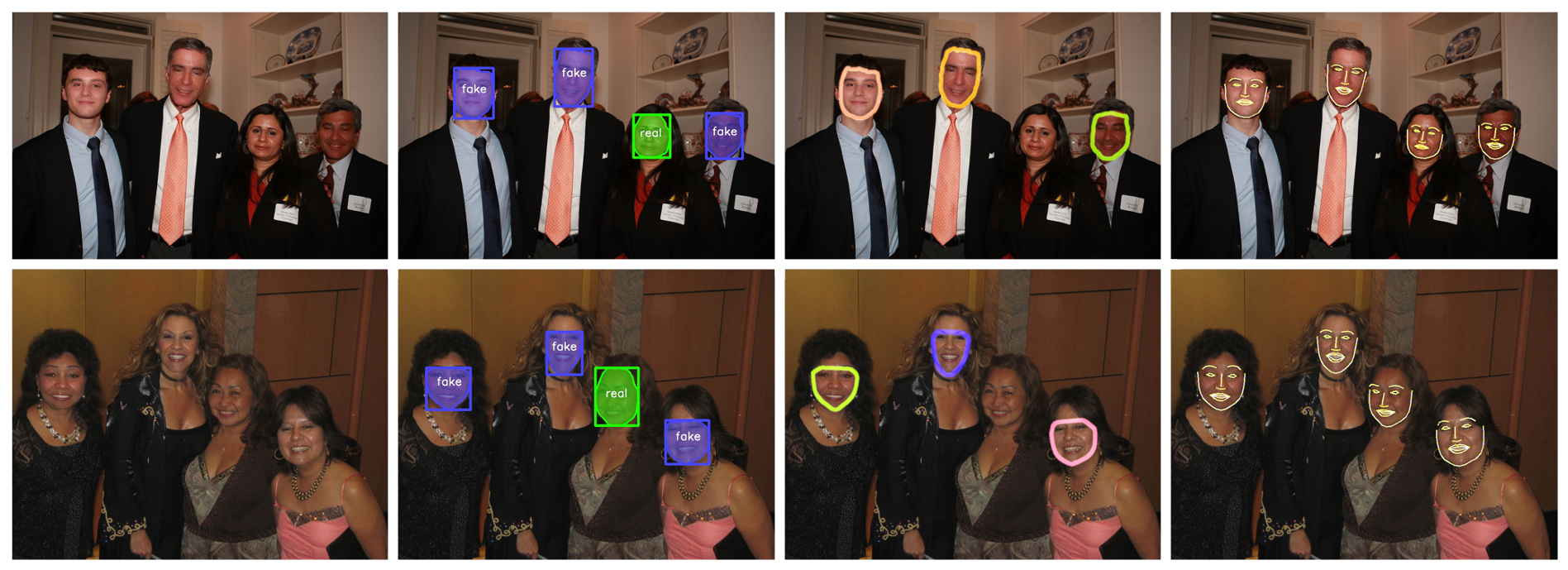}
	\caption{Face-wise multi-task ground truth in OpenForensics dataset (best viewed online in color with zoom-in). From left to right, original images followed by overlaid ground truth bounding box and segmentation mask, forgery boundary, and general facial landmarks.}
	\label{fig:multi_task_annotation}
	\vspace{-5mm}
\end{figure}

Conventional face forgery recognition methods~\cite{Afchar-WIFS2018, nhhuy-ICASSP2019, nhhuy-BTAS2019} require the input of given face regions. Therefore, they can process only one face at a time, and processing multiple faces sequentially is time-consuming. Moreover, their performance greatly depends on the accuracy of the independent face detection method used. Given that these methods have been evaluated only in laboratory environments using images with a simple background and a single clear front face~\cite{Korshunov-2018, Yang-ICASSP2019}, they are not ready for deployment in the real world, where the contexts are much more diverse and challenging than simple staged scenarios. 

\begin{table*}[t!]
\vspace{-5mm}
\caption{Basic information about deepfake datasets. “Cls.”, “Det.” and “Seg.” stand for classification, detection, and segmentation, respectively. Pristine scenarios are originally collected images/videos used to generate fake data. Unique fake scenarios are fake images/videos ignoring perturbations. \highlight{Released scenarios are number of real/fake (or both) images/videos publicly released by authors.}}
\label{table:deepfake_datasets}
\resizebox{1\linewidth}{!}{
\begin{tabular}{l|c|c|c|c|c|c|c|c|c|c}
\hline
\textbf{Dataset} & \textbf{Year} & \textbf{Task} & \textbf{GT Type} & \textbf{Fake Identity} & \begin{tabular}[c]{@{}c@{}}\textbf{\#Face}\\\textbf{Per Image}\end{tabular} & \begin{tabular}[c]{@{}c@{}}\textbf{Face}\\\textbf{Occlusion}\end{tabular} & \begin{tabular}[c]{@{}c@{}}\textbf{\#Pristine}\\\textbf{Scenario}\end{tabular} &
\begin{tabular}[c]{@{}c@{}}\textbf{\#Unique Fake}\\\textbf{Scenario}\end{tabular} &\begin{tabular}[c]{@{}c@{}}\textbf{\#Released}\\\textbf{Scenario}\end{tabular} & \begin{tabular}[c]{@{}c@{}}\textbf{Data}\\\textbf{Augmentation}\end{tabular} \\
\hline
DF-TIMIT~\cite{Korshunov-2018} & 2018 &  Cls. & Image label  & Other videos & 1 & \xmark & 320 & 320 & 640 & \xmark \\
UADFV~\cite{Yang-ICASSP2019} & 2019 &  Cls. & Image label  & Other videos & 1 & \xmark & 49 & 49 & 98 & \xmark \\
FaceForensics++~\cite{Rossler-ICCV2019} & 2019 & Cls. & Image label  & Other videos & 1 & \xmark & 1,000 & 4,000 & 5,000 &  \xmark \\
Google DFD~\cite{google_dfd-2019} & 2019 &  Cls. & Image label  & Other videos & 1 & \xmark & 363 & 3,068 & 3,431 & \xmark \\
Facebook DFDC~\cite{Dolhansky-2020}  & 2020 & Cls. & Image label  & Other videos & 1 & \xmark & 48,190 & 104,500 & 128,154 & \cmark \\
Celeb-DF~\cite{Yuezun-CVPR2020} & 2020 & Cls. & Image label  & Other videos & 1 & \xmark & 590 & 5,639 & 6,229 & \xmark \\
DeeperForensics~\cite{Jiang-CVPR2020} & 2020 & Cls. & Image label & Hired actors & 1 & \xmark & 1,000 & 1,000 & 10,000 & \cmark \\
WildDeepfake~\cite{Bojia-MM2020} & 2020 & Cls. & Image label & N/A & 1 & \xmark & 0 & 707 & N/A & \xmark \\
\rowcolor{lightgray} OpenForensics & 2021 & Det. / Seg. & BBox/Mask & GAN & $>1$ & \cmark & 45,473 & 70,325 & 115,325 & \cmark \\
\hline
\end{tabular}
}
\vspace{-4mm}
\end{table*}

It has thus become essential to develop methods that can effectively process multiple faces simultaneously from an input image. To our best knowledge, no methods have been proposed for face forgery detection and segmentation officially. We attribute this partially to the lack of a large-scale dataset for training and testing. To encourage more studies in this field, we present four contributions in this paper. 

First, we present a comprehensive study on tasks related to massive face forgery in-the-wild. Particularly, we introduce two new tasks: \textit{multi-face forgery detection and segmentation in-the-wild}. This is the first formal exploration of these tasks to the best of our knowledge. Previous work has explored only single-face forgery recognition.

Second, we propose generating an infinite number of fake individual identities using GAN models for non-target face-swapping without repeatedly training a deepfake AE. Our proposed forgery workflow reduces the cost of synthesizing fake data.

Third, using the proposed forgery workflow, we introduce a novel image dataset to support the development of multi-face forgery detection and segmentation tasks. Our newly constructed \textit{OpenForensics dataset is the first large-scale dataset designed for these tasks}. It consists of 115K \highlight{unrestricted} images with 334K human faces. \highlight{Unlike existing datasets, ours contains various backgrounds and multiple people of various ages, genders, poses, positions, and face occlusions.} All images have \textit{face-wise rich annotations} supporting multiple tasks, such as forgery category, bounding box, segmentation mask, forgery boundary, and general facial landmarks (see Figs.~\ref{fig:examples} and \ref{fig:multi_task_annotation}). The dataset can thus support not only multi-face forgery detection and segmentation tasks but also conventional tasks involving the general human face.  

Fourth, we present a benchmark suite to facilitate the evaluation and advancement of these tasks. We conducted an extensive evaluation and in-depth analysis of state-of-the-art instance detection and segmentation models in various scenarios.

The whole dataset, evaluation toolkit, and trained models will be freely available on our project page\footnote{\url{https://sites.google.com/view/ltnghia/research/openforensics}}.


\section{Related Work}
\label{sec:related_work}

\subsection{Existing Forensic Datasets}

Table~\ref{table:deepfake_datasets} summarizes basic information about existing forensic datasets. The DF-TIMIT dataset~\cite{Korshunov-2018} has 640 fake videos crafted from Vid-TIMIT dataset~\cite{Sanderson-ICB2009} using Faceswap-GAN~\cite{faceswap-gan}. The UADFV dataset~\cite{Yang-ICASSP2019} consists of 98 videos, half of which are fake, created using FakeAPP~\cite{faceapp}. The FaceForensics++ dataset~\cite{Rossler-ICCV2019} contains 1000 pristine videos from YouTube and 4000 synthetic videos manipulated using deepfake methods~\cite{deepfakes, Thies-CVPR2016, faceswap, Thies-TG2019}. The Google DFD dataset~\cite{google_dfd-2019} includes 3068 fake videos. The Facebook DFDC dataset~\cite{Dolhansky-2020} contains 128K original and manipulated videos created using various deepfake and augmentation methods~\cite{Ivan-2020, Huang-ECCV2012, Egor-ICCV2019, Nirkin-ICCV2019, Karras-CVPR2019}. The Celeb-DF dataset~\cite{Yuezun-CVPR2020} comprises YouTube celebrity videos and 5,639 fake videos. The DeeperForensics dataset~\cite{Jiang-CVPR2020} consists of 10K manipulated videos using a deepfake VAE and augmentations on 1000 original videos in the FaceForensics++ dataset. The WildDeepfake dataset~\cite{Bojia-MM2020} contains face sequences extracted from 707 deepfake videos collected from the Internet. As shown in Table~\ref{table:deepfake_datasets}, our OpenForensics is the first dataset designed for face forgery detection and segmentation.

\highlight{Existing forensic datasets were created by dividing long videos into short ones, leading to that even pristine videos have the same background. Subsequent synthesizing many fake videos from one pristine video resulted in lots of similar backgrounds.} Deep models trained on the existing datasets may not generalize well to the real world due to the \highlight{repeated background}. 
In contrast, our large-scale image dataset contains diverse backgrounds. Inspired by the work of Dolhansky \etal~\cite{Dolhansky-2020} and Jiang \etal~\cite{Jiang-CVPR2020}, we systematically applied a mixture of perturbations to raw manipulated images to imitate real-world scenarios. With the existing datasets, a deepfake model needs to be trained on each pair of videos to swap human identities, yielding a considerable number of models requiring training. In contrast, a massive number of fake faces in our dataset are synthesized by GAN without repeatedly re-training deepfake models. While existing datasets were developed for only the single-face forgery classification task, our dataset is the first one designed for multi-face forgery detection and segmentation tasks, which require more annotation than the classification task. Our dataset can also be utilized for various general face-related tasks.

\begin{figure}[t!]
	\centering
	\includegraphics[width=1\linewidth]{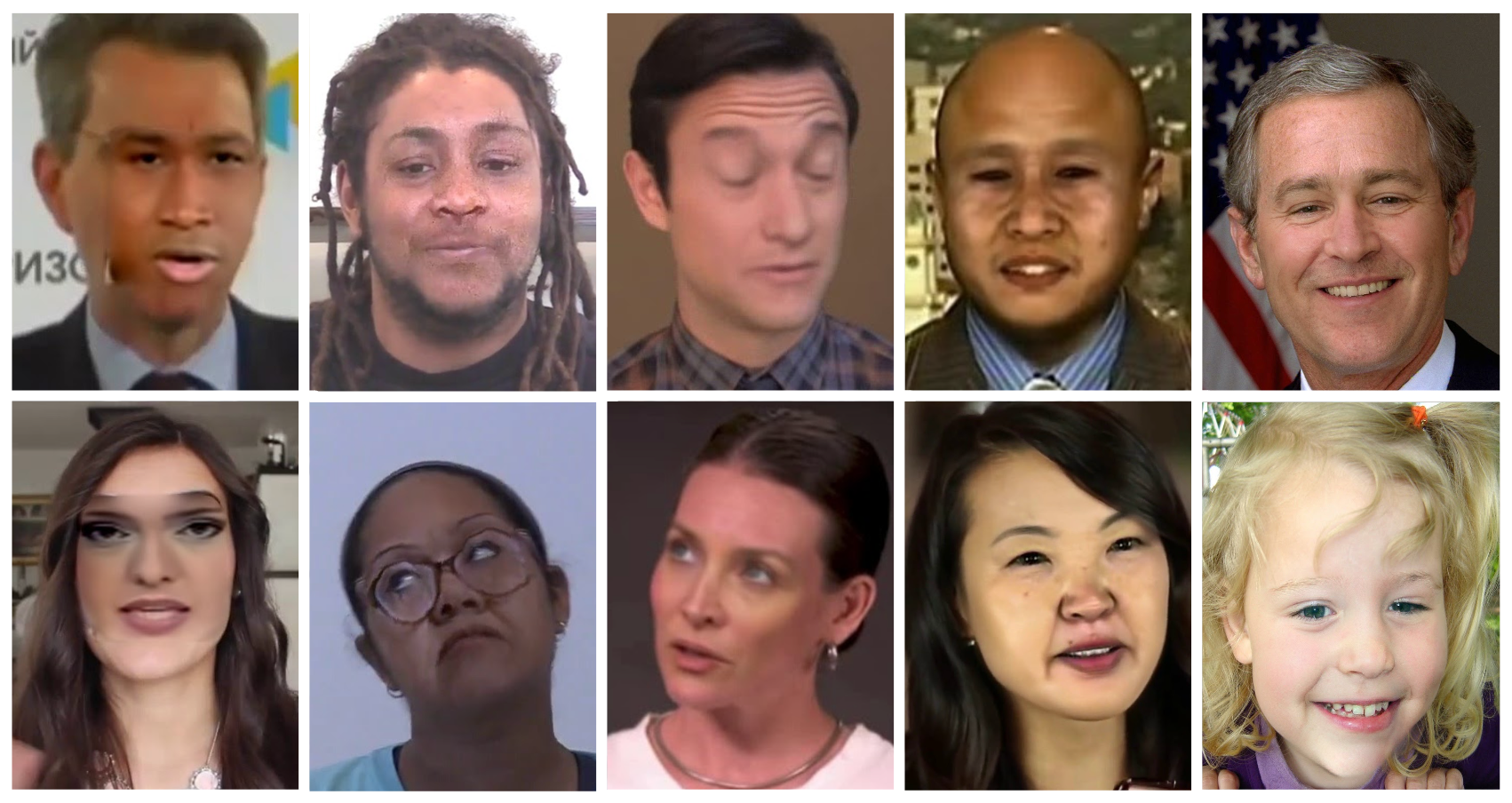}
	\caption{Visual artifacts of forged faces in datasets. From left to right, FaceForensics++~\cite{Rossler-ICCV2019}, DFDC~\cite{Dolhansky-2020}, DeeperForensics~\cite{Jiang-CVPR2020}, Celeb-DF~\cite{Yuezun-CVPR2020}, and our OpenForensics. Faces generated in our dataset have the highest resolution and best quality.}
	\label{fig:synthesized_face}
	\vspace{-5mm}
\end{figure}

\subsection{Face Manipulation and Generation}

A number of deepfake open-source techniques for swapping human faces have been released~\cite{faceswap, deepfakes, faceswap-gan}. These techniques have gradually evolved from using hand-crafted features~\cite{faceswap} to using deep learning by training AE architectures~\cite{deepfakes} and GAN models~\cite{faceswap-gan}~\cite{Lingzhi-CVPR2020} to achieve realism. Facial reenactment techniques have been developed for transferring expressions~\cite{Thies-CVPR2016, Thies-TG2019, Nirkin-ICCV2019}. Different techniques such as 3D reconstruction~\cite{Thies-CVPR2016} and neural textures~\cite{Thies-TG2019} were used to preserve the target skin color and lighting conditions. Boundary latent space~\cite{Wayne-ECCV2018} and disentangle shape~\cite{Zhixin-ECCV2018} were combined with AE models to morph expressions. In addition to transferring expressions, the head pose can be controlled by using a recurrent neural network to enhance naturalness~\cite{Nirkin-ICCV2019} by using different modalities~\cite{Wiles-ECCV2018} and by using human interpretable attributes and actions~\cite{Soumya-WACV2020}. 

Subsequently proposed techniques for face synthesis use deep learning. They generally use GAN for facial attribute translation~\cite{Choi-CVPR2018, Choi-CVPR2020, Karras-CVPR2019, Karras-CVPR2020}, for identity-attribute combination~\cite{Bao-CVPR2018}, for identified characteristics removal~\cite{Maximov-CVPR2020}, and for interactive semantic manipulation~\cite{ChengHan-CVPR2020, Zhu-CVPR2020}. Facial disentangled features are being interpreted in different latent spaces, resulting in more precise control of attribute manipulation in face editing~\cite{Karras-CVPR2019, Karras-CVPR2020, Shen-CVPR2020, Pidhorskyi-CVPR2020}.

Existing deepfake methods require face pairs for specific training, meaning that the cost of training is very high. Training requires sequences of images; thus, these methods are practical only for videos, and the generated faces usually have low-resolution. Although existing face synthesis methods can generate high-quality faces, the synthesized faces are oriented to the front and are not consistent with the original faces if the original faces are not close to the distribution of the training data. We combine these two approaches to generate an infinite number of fake human identities without repeatedly training the AEs. We achieve this by transforming GAN-based high-quality synthesized faces into original poses.

\begin{table}[t!]
\caption{Scale of object detection/segmentation datasets.}
\label{table:datasets}
\resizebox{1\linewidth}{!}{
\begin{tabular}{l|c|l|c|l}
\hline
\multicolumn{1}{c}{\textbf{Dataset}} & \multicolumn{1}{|c}{\textbf{Year}} & \multicolumn{1}{|c}{\textbf{\begin{tabular}[c]{@{}c@{}}Object Type\end{tabular}}} & \textbf{\#Annotated Images} & \textbf{Ground-Truth Type} \\
\hline
COCO~\cite{Lin-ECCV2014} & 2014 & General object & 200,000 & Coarse mask \\
CityScapes~\cite{Cordts-CVPR2016} & 2016 & Road object & 25,000 & Coarse\&Fine mask \\
WiderFace~\cite{Shuo-CVPR2016} & 2016 & Human face & 32,200 & Bounding box \\
SESIV~\cite{ltnghia-WACV2019} & 2019 & Salient object & 5,700 & Fine mask \\
ADV~\cite{ltnghia-IV2020} & 2020 & Accident object & 10,000 & Fine mask \\
CAMO++~\cite{ltnghia-2021} & 2021 & Camouflaged object & 5,500 & Fine mask \\
\rowcolor{lightgray} OpenForensics & 2021 & Forged face & 115,325 & Fine mask \\
\hline
\end{tabular}
}
\end{table}

\begin{table}[t!]
\caption{Image distribution in OpenForensics dataset.}
\label{table:dataset_splits}
\footnotesize
\resizebox{1\linewidth}{!}{
\begin{tabular}{l|r|r|r|r}
\hline
\textbf{Subset} & \textbf{\#Images} & \textbf{\#Faces} & \textbf{\#Real Faces} & \textbf{\#Forged Faces} \\
\hline
Training & 44,122 & 151,364 & 85,392 & 65,972 \\
Validation & 7,308 & 15,352 & 4,786 & 10,566 \\
Test-Development & 18,895 & 49,750 & 21,071 & 28,670 \\
Test-Challenge & 45,000 & 117,670 & 49,218 & 68,452  \\
\rowcolor{lightgray} Total & 115,325 & 334,136 & 160,67 & 173,660  \\
\hline
\end{tabular}
}
\vspace{-5mm}
\end{table}

\subsection{Face Forgery Classification}

Researchers have been investigating the problem of face forgery classification, which is generally regarded as merely a binary classification problem (real/fake). The research task is also called `deepfake detection,' but the term `detection' may lead to a misunderstanding of the fundamental task of \textit{object detection}. Early methods exploited inconsistencies created by visual artifacts in deepfake images and videos by analyzing biological clues such as eye blinking~\cite{Li-WIFS2018}, head pose~\cite{Yang-ICASSP2019}, skin texture~\cite{Liu-CVPR2020}, and iris and teeth color~\cite{Matern-WACVW2019}. A few works investigated artifacts in affine face warping~\cite{Yuezun-CVPRW2019} or in the blending boundary~\cite{Li-CVPR2020} to distinguish real and fake faces. Most current methods are data-driven, directly training deep networks on real and fake images and videos~\cite{Afchar-WIFS2018, nhhuy-ICASSP2019, Rossler-ICCV2019, nhhuy-BTAS2019, Zhou-CVPRW2017, Wang-IJCAI2020}. They do not rely on specific artifacts.

Existing face forgery classification approaches do not have a face localization ability. They can work only on a single cropped face; thus, their performance relies heavily on independent face detection performed as pre-processing. To the best of our knowledge, ours is the first work addressing multi-face detection and segmentation in-the-wild.


\section{Large-Scale OpenForensics Dataset}
\label{sec:dataset}

The emergence of new tasks and datasets has led to rapid progress in human research areas~\cite{Shuo-CVPR2016, Deng-CVPR2020, vindrcxr, Haghighi-MICCAI2020, Guo-ECCV2020}. However, research on human forgery prevention is only now beginning, and the field is still immature with work only on the face forgery classification task. With this in mind, our goal is to study and develop a dataset that will support challenging new forgery research tasks in both the computer vision and forensic communities.

\subsection{Dataset Construction}

\begin{figure}[t!]
\vspace{-5mm}
	\centering
	\includegraphics[width=1\linewidth]{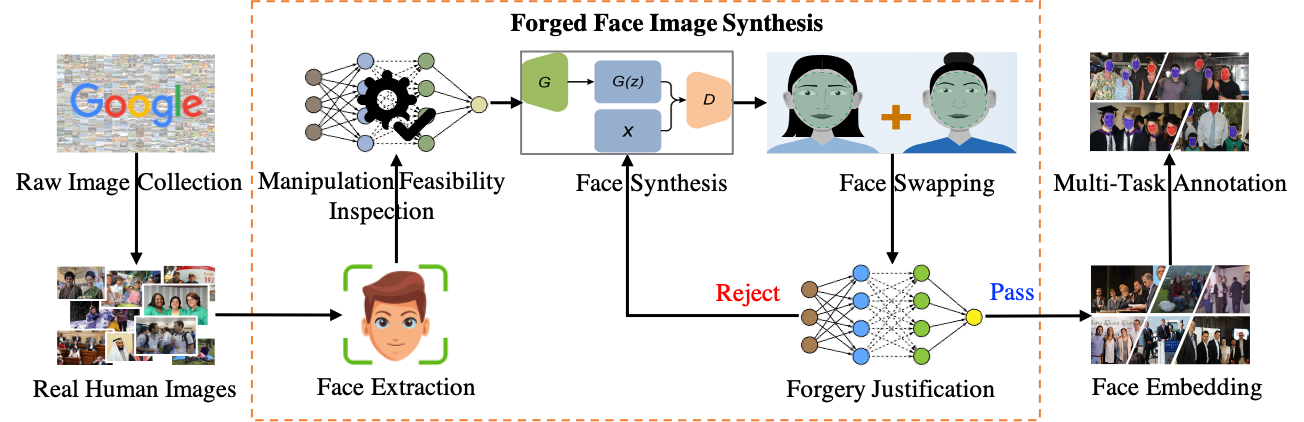}
	\caption{Dataset construction workflow: 1) collect raw images and manually select real face images; 2) synthesize forged face images (for each original extracted face, new identities are repeatedly generated until swapped faces can spoof our simple classifier); 3) perform face-wise multi-task annotation.}
	\label{fig:dataset_construction2}
\end{figure}

As shown in Fig.~\ref{fig:dataset_construction2}, the dataset construction workflow includes three main steps: real human image collection, forged face image synthesis, and multi-task annotation.

\subsubsection{Real Human Image Collection}

We collected raw images from Google Open Images~\cite{Kuznetsova-IJCV2020} and removed images without people. Images consisting of unreal human faces (\eg, images on money and in books, magazines, cartoons, and sketches) or human-like objects (\eg, dolls, robots, and sculptures) were also removed. We ended up with 45,473 images, which were used as pristine data.

\subsubsection{Forged Face Image Synthesis}

Figure \ref{fig:dataset_construction2} shows an overview of the process used to synthesize forged face images. First, all faces in the real human images are extracted and checked in the manipulation feasibility inspection module to see whether they can be manipulated. This is done using various conditions (\eg, face size, image quality, and blurring) and a random manipulation probability. If manipulation is feasible, the image undergoes a cyclical process. Inspired by GAN-based face synthesis~\cite{Choi-CVPR2020, Karras-CVPR2020}, we first extract the facial identity latent vector and modify it using random values. The modified latent vector is then fed into GAN models~\cite{Shen-CVPR2020, Pidhorskyi-CVPR2020} to generate a new face. The synthesized face is subsequently transformed into an original pose. Feasible manipulation regions in the synthesized face (\eg, regions inside facial landmarks or the entire face) are extracted and blended into the original face using Poisson blending~\cite{Patrick-SIGGRAPH2003} and a color adaptation algorithm in the face-swapping module, with the final result being a new identity. The new identity image is then tested to determine whether it can spoof a simple classifier (\ie, XceptionNet~\cite{Chollet-CVPR2017}) in the forgery justification module, which is trained to distinguish real and fake identities. Those for which spoofing is successful are overlaid onto the original image. The others are discarded, and new faces are generated. \highlight{We provide detailed implementation and training of networks in the supplementary material.}

\highlight{Our synthesis workflow features the ability to synthesize an unlimited number of fake identities at low cost for non-target face-swapping without paired training. Meanwhile, other deepfake methods use a limited number of fake identities extracted from videos and perform paired training using deep models for target face-swapping. They thus require much time and resources to synsthesize datasets.} Our synthesis approach also overcomes the limitations of existing approaches. Existing approaches~\cite{Rossler-ICCV2019, Dolhansky-2020, Jiang-CVPR2020} generate low-resolution faces (typically less than $256 \times 256$ pixels) while our approach generates faces with \textit{higher resolution (\ie, $512 \times 512$ pixels) and better visual quality} (cf. Fig.~\ref{fig:synthesized_face}). Our use of Poisson blending~\cite{Patrick-SIGGRAPH2003} and a color adaptation algorithm to reduce the color mismatch between the synthesized and original face (Fig.~\ref{fig:synthesized_face}) \textit{enhances the naturalness of the forged faces}. We also \textit{improve the smoothness of the blending mask} by extracting 68 facial landmark points and training face segmentation models, resulting in fine boundaries and complete facial coverage (see Fig.~\ref{fig:multi_task_annotation} for different blending masks). The blending masks used to create existing datasets were either rectangular or rough convex hulls between the eyebrows and lower lip, resulting in incomplete facial coverage or visible boundaries (cf. Fig~\ref{fig:synthesized_face}). 


Finally, we randomly split the accepted images into separate training, validation, and test-development sets (ratio of 60:10:30). Table \ref{table:dataset_splits} shows the distribution of images and faces in our newly constructed OpenForensics dataset.

\begin{figure}[t!]
\vspace{-5mm}
	\centering
	\includegraphics[width=1\linewidth]{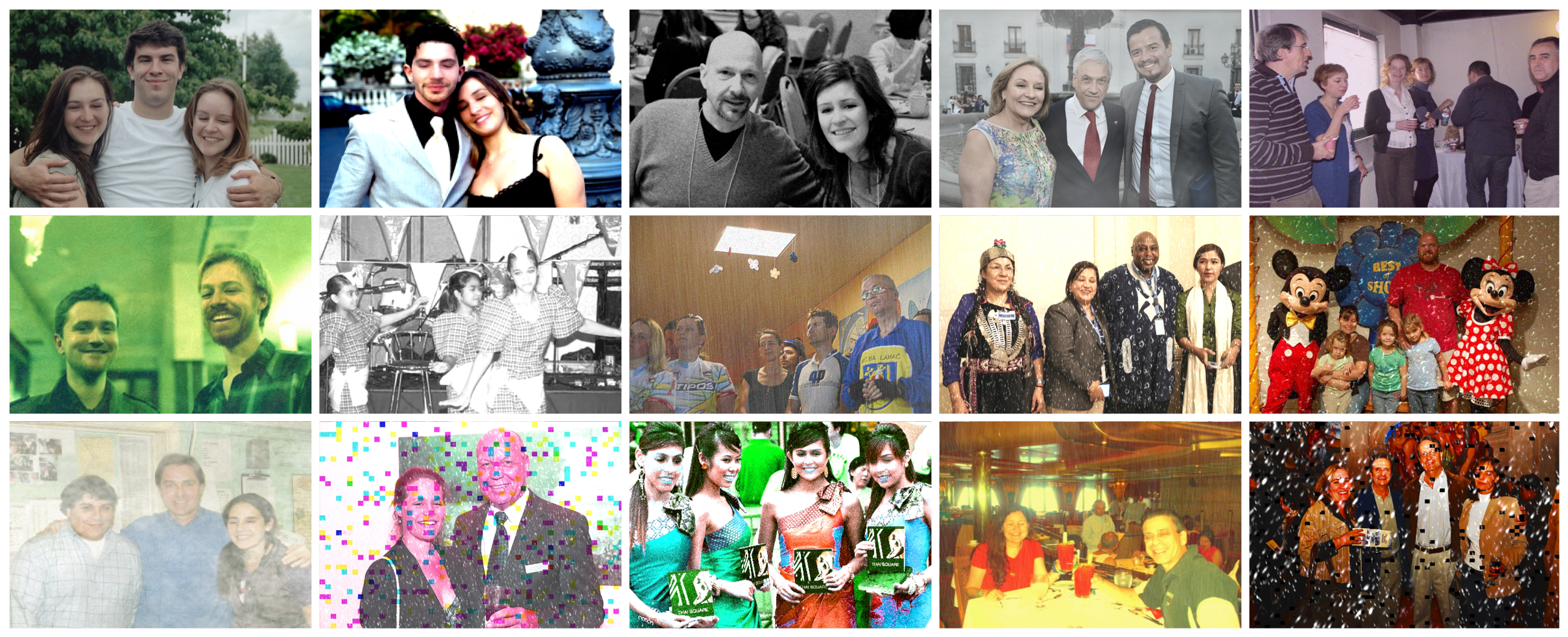}
	\caption{Example images in test-challenge set (three levels: easy, medium, and hard from top to bottom). Each image contains at least one forged face. See supplementary material for overlaid ground truth.}
	\label{fig:augmentations}
	\vspace{-5mm}
\end{figure}

\subsubsection{Challenging Scenario Augmentation}

To enhance the challenges posed by our OpenForensics dataset for real-world face forgery detection and segmentation, we applied various perturbations to better simulate contexts in natural scenes, resulting in a test-challenge subset. Various augmented operators are divided into overarching groups.
\begin{itemize}
	\item Color manipulation: Hue change, saturation change, brightness change, histogram adjustment, contrast addition, grayscale conversion.
	\item Edge manipulation: edge detection and alteration.
	\item Block-wise distortion: color grouping, color pooling, color quantization, and pixelation. 
	\item Image corruption: elastic deformation, jigsaw distortion, JPEG compression, noise addition, and dropout.
	\item Convolution mask transformation: Gaussian blurring, motion blurring, sharpening, and embossing.
	\item External effect: fog, cloud, sun, frost, snow, and rain.
\end{itemize}

These augmentations are divided into three intensity levels (\ie, easy, medium, and hard) to ensure diverse scenarios. For each level, random-type augmentation is applied separately or as a mixture, resulting in 45,000 images. Example images in the test-challenge set are shown in Fig.~\ref{fig:augmentations}. 

\subsection{Dataset Description}

\textbf{Task Diversity.} Existing deepfake datasets~\cite{Rossler-ICCV2019, Dolhansky-2020, Jiang-CVPR2020, Yuezun-CVPR2020} focus exclusively on video-wise labels for classification. In contrast, we aim to exploit the face-wise ground truth, which requires much more annotation effort, to advance further forgery analysis. Each face was labeled with various ground-truths such as forgery category (real/fake), bounding box, segmentation mask, forgery boundary, and facial landmarks (cf. Fig.~\ref{fig:multi_task_annotation}). Our rich annotation can be utilized for various tasks and even multi-task learning.

\textbf{Dataset Size.} OpenForensics is one of the largest detection and segmentation datasets (cf. Table~\ref{table:datasets}) and is large enough to train and evaluate deep networks. This should encourage more research in this field. 

\textbf{Diverse Scenarios.} Existing datasets~\cite{Rossler-ICCV2019, Dolhansky-2020, Jiang-CVPR2020, Yuezun-CVPR2020} were released as short videos. Although they contain a vast number of images, frames in a short video are similar and do not contribute much to the training of deep networks. With these datasets, data sampling is usually used for training deep networks to avoid overfitting and to reduce training time. We define similar frames in a short video as a `scenario' and assert that training using a diversity of scenarios helps to make deep networks more effective. Table~\ref{table:deepfake_datasets} shows that the OpenForensics dataset is an order of magnitude larger than existing datasets in terms of the number of scenarios, with only slightly fewer than in the DFDC dataset.

\begin{figure*}[t!]
    \vspace{-5mm}
    \centering
    \begin{tabularx}{\linewidth}{*{6}{X}}
        \hfill\includegraphics[width=0.87\linewidth]{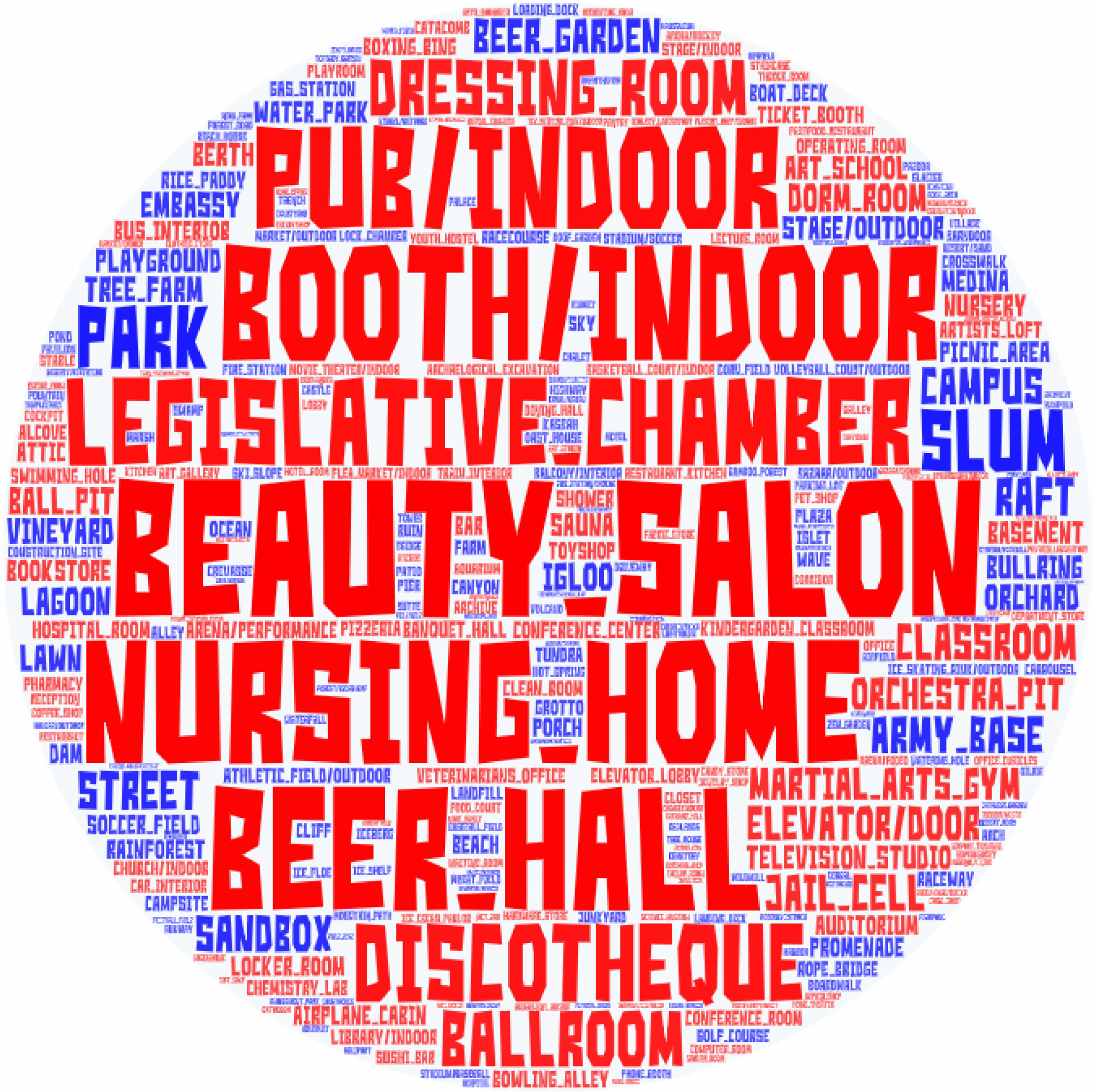}\hspace*{\fill} & 
        \hfill\includegraphics[width=0.87\linewidth]{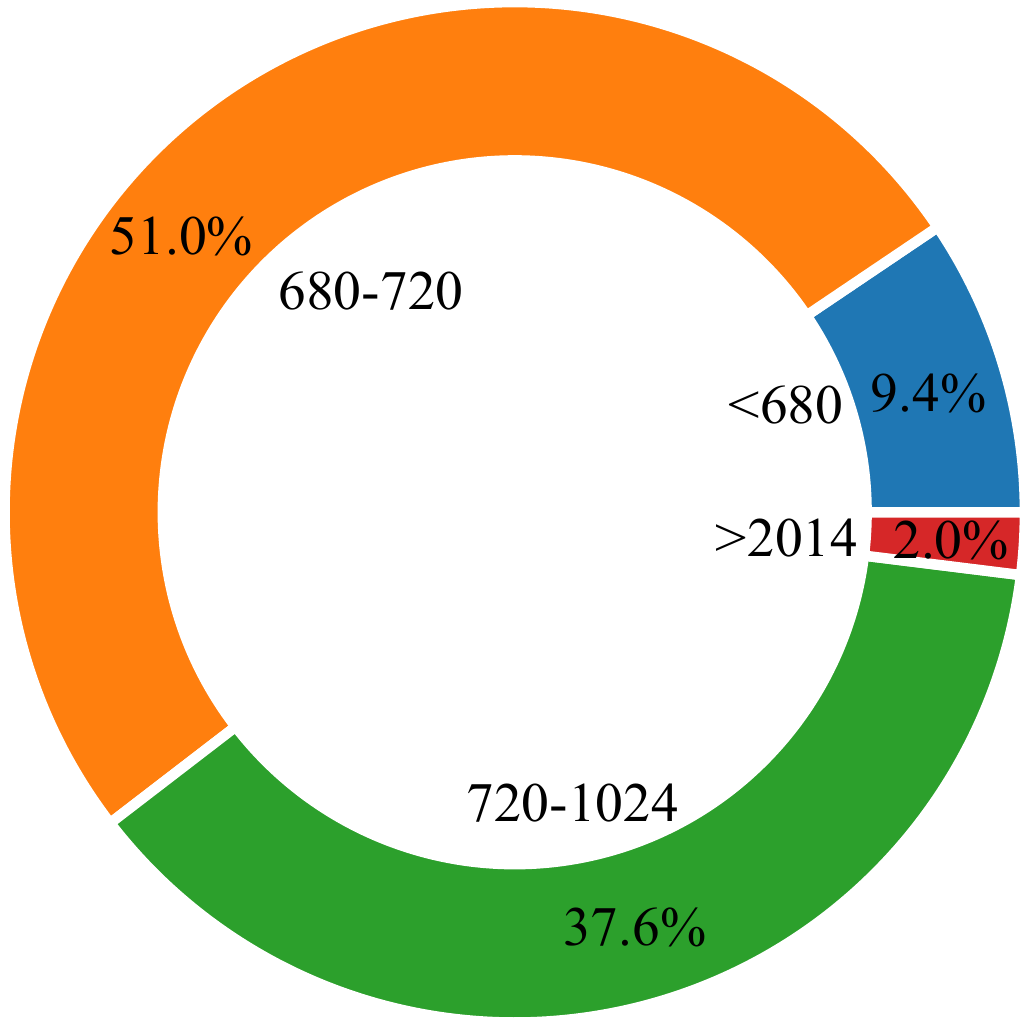}\hspace*{\fill} &
        \hfill\includegraphics[width=0.87\linewidth]{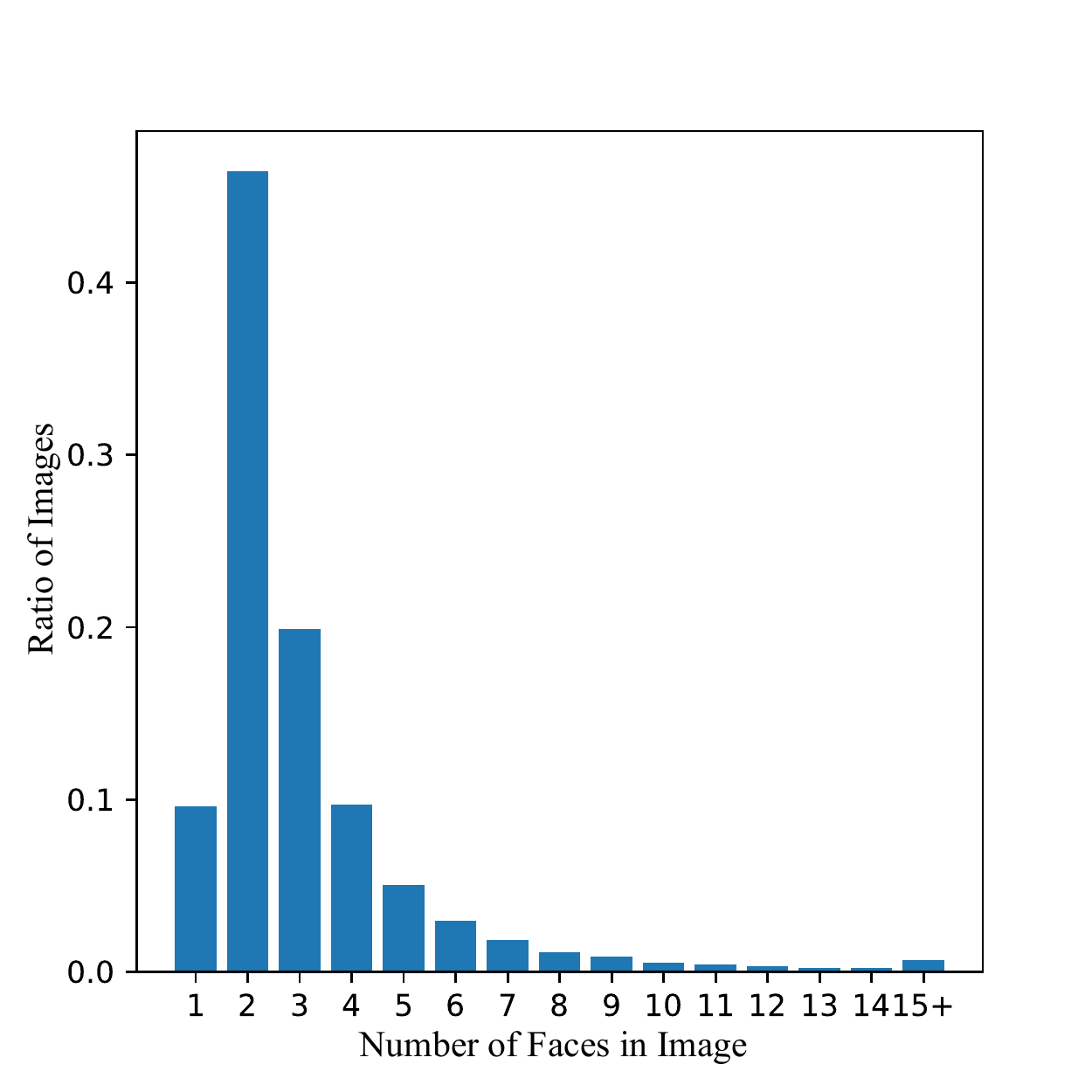}\hspace*{\fill} &
        \hfill\includegraphics[width=1\linewidth]{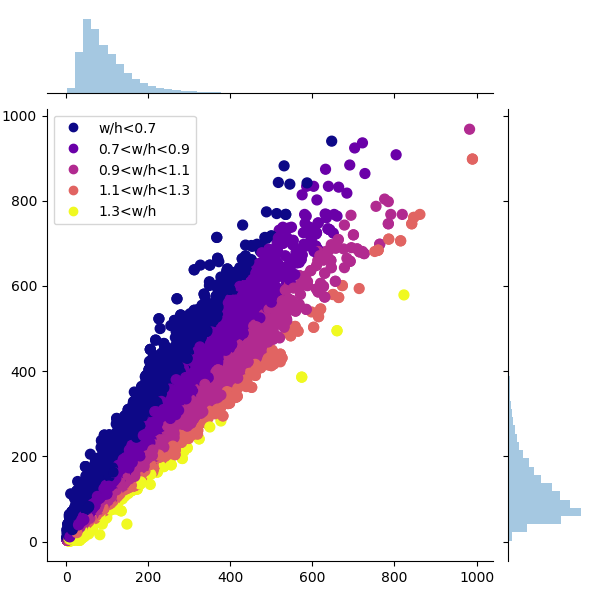}\hspace*{\fill} &
        \hfill\includegraphics[width=0.92\linewidth]{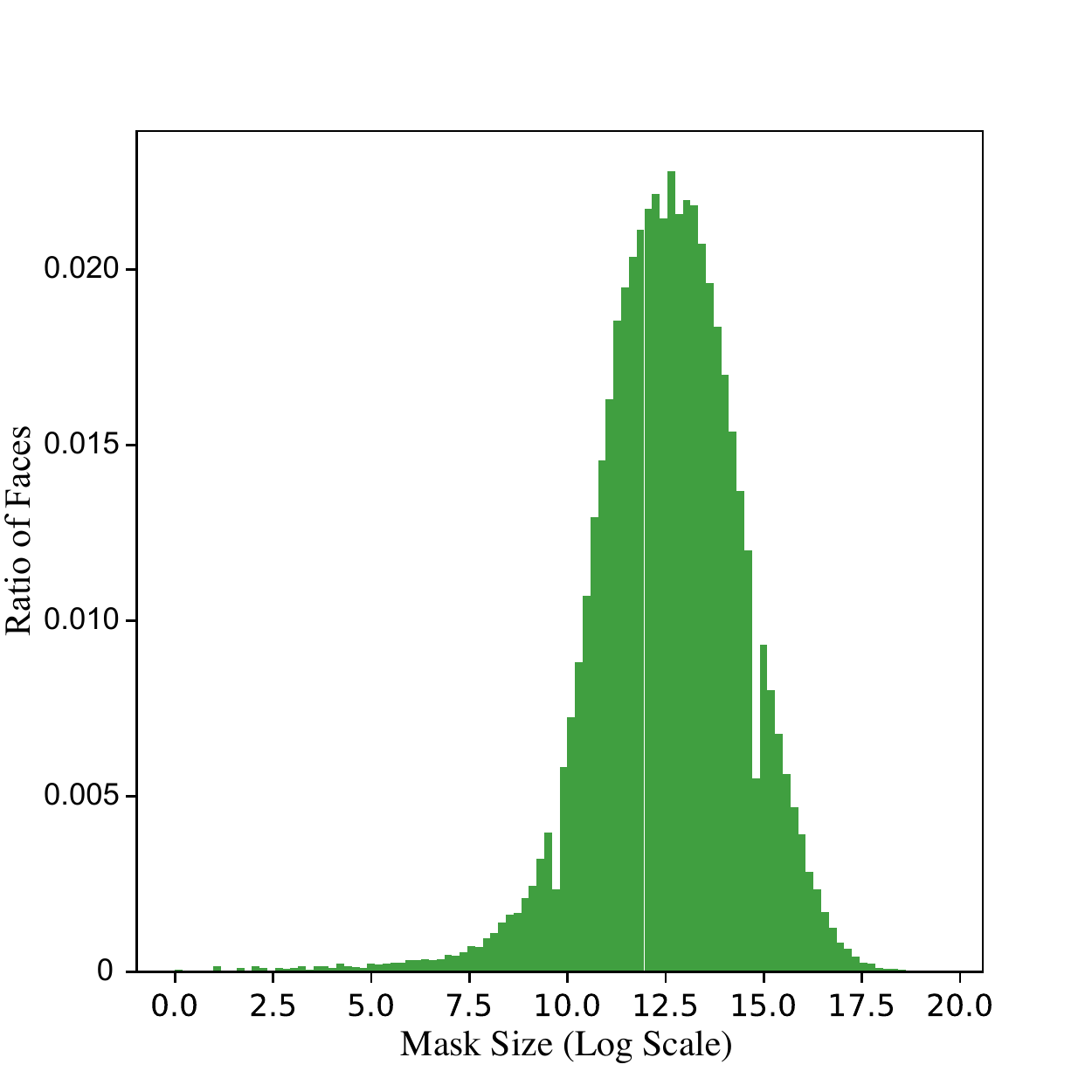}\hspace*{\fill} & 
        \hfill\includegraphics[width=0.92\linewidth]{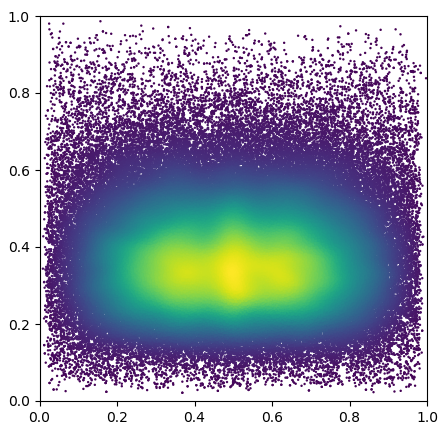}\hspace*{\fill} \\
        \centering \scriptsize{a) Scene word cloud} &
        \centering \scriptsize{b) Image resolution} &
        \centering \scriptsize{c) Faces per image} &
        \centering \scriptsize{d) Bounding box size} &
        \centering \scriptsize{e) Mask size} &
        \centering \scriptsize{f) Face centroid}
    \end{tabularx}
    \caption{Distributions in OpenForensics dataset (best viewed online in color with zoom-in). In image scene distribution, red represents indoor scenes and blue represents outdoor scenes (percent of indoor scenes is 63.7\%). There are 2.9 faces per image on average.}
    \label{fig:data_distribution}
    \vspace{-3mm}
\end{figure*}

\textbf{Image Scene.} Existing deepfake datasets~\cite{Rossler-ICCV2019, Yuezun-CVPR2020} contain limited types of image scenes, such as indoor scenes and television scenes. In contrast, the OpenForensics dataset contains various types of scenes. We computed scenes using a pre-trained model on the large-scale Places2 dataset~\cite{Zhou-TPAMI2017}. Figure \ref{fig:data_distribution}(a) shows the distribution as a word cloud, with the various outdoor scenes accounting for 36.3\% of the images.

\textbf{Image Resolution.} Figure \ref{fig:data_distribution}(b) shows the distribution of image resolutions in the OpenForensics dataset. The large number of high-resolution images, which provide more face boundary details for model training, results in better performance.

\textbf{Multiple Faces Per Image.} Existing deepfake datasets~\cite{Rossler-ICCV2019, Dolhansky-2020, Jiang-CVPR2020, Yuezun-CVPR2020} mostly have only one face per image. In contrast, the OpenForensics dataset has multiple faces per image (2.9 on average). Figure \ref{fig:data_distribution}(c) shows the distribution.

\textbf{Face Characteristics.} Figures \ref{fig:data_distribution}(d and e) show the distribution of faces in the OpenForensics dataset by bounding box size and mask size (\ie, number of pixels covering face). OpenForensics contains faces of various sizes, from tiny to large. The distribution of face centroids in Fig.~\ref{fig:data_distribution}(f) shows that the faces tend to be near the image center. In addition, the ratio of male and female faces is 50:50, and there is a diversity of ages. More details are provided in the supplementary material.

\textbf{Data Augmentation.} Deep models trained on existing deepfake datasets may not perform well in the real world due to overfitting caused by image similarity in the training data. Although strong deep models have obtained very high accuracy~\cite{nhhuy-ICASSP2019, Li-CVPR2020}, even near 100\%, they may easily fail in the real world if they do not share a close distribution with the training dataset. To simulate real-world contexts in the OpenForensics dataset, diverse perturbations were used to improve scenario diversity so as to better imitate real-world data distributions. Improvements have been made to a couple of existing datasets by using simple perturbations, which have increased their size. For instance, the DFDC dataset~\cite{Dolhansky-2020} and DeeperForensics dataset~\cite{Jiang-CVPR2020} have been improved by applying geometric and color transforms, adding noise, blurring, and overlaying objects.


\subsection{User Study}

To evaluate the visual quality of the images in the OpenForensics dataset and human performance in face forgery detection, we conducted a user study with \highlight{200} participants, \highlight{80} of whom are \highlight{experts, who can provide knowledgeable opinions due to their researching deepfakes}. 
The study results can fairly reflect the performance of both experts and non-experts.

The study was conducted on the OpenForensics dataset and four existing deepfake datasets: FaceForensics++~\cite{Rossler-ICCV2019}, DFDC~\cite{Dolhansky-2020}, Celeb-DF~\cite{Yuezun-CVPR2020} and DeeperForensics~\cite{Jiang-CVPR2020}. For each dataset, we randomly selected 600 images and prepared a virtual platform for the participants. 

\begin{figure}[t!]
    \vspace{-5mm}
	\centering
	\includegraphics[width=1\linewidth]{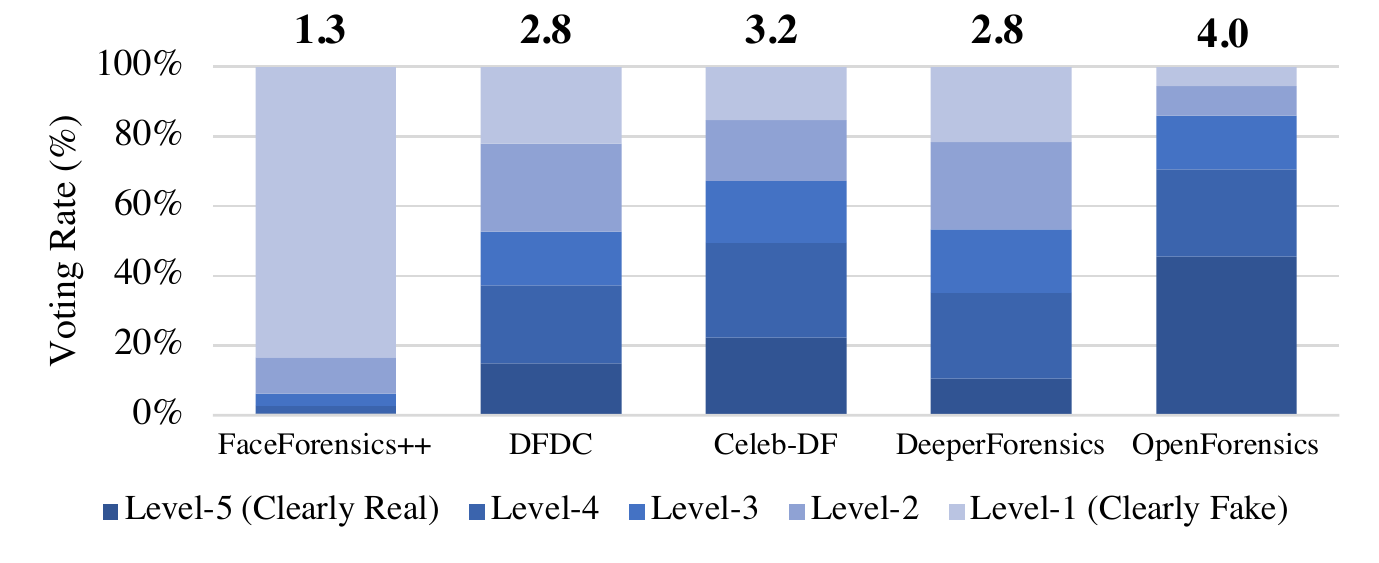}
	\caption{Distributions of image realism scores for five compared datasets. Mean opinion scores (MOS) are shown at top of bars. OpenForensics dataset achieved highest MOS and had highest percentage of level-5 scores.}
	\label{fig:study_realistic}
	\vspace{-3mm}
\end{figure}

\begin{figure}[t!]
	\centering
	\includegraphics[width=1\linewidth]{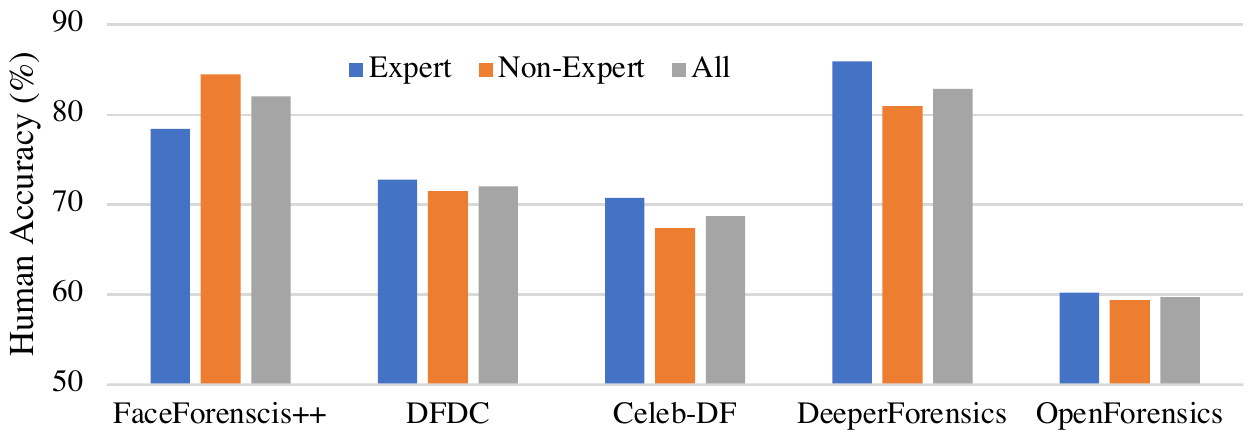}
	\caption{Human accuracy in face forgery classification. Images in OpenForensics dataset were most effective in spoofing both experts and non-experts.}
	\label{fig:subjective_test_detection_human_accuracy}
	\vspace{-3mm}
\end{figure}

\highlight{We argue that participants can quickly see that a face is fake if they see two similar images but different people, leading to unfair comparison with existing datasets. In addition, the forgery identification may becomes difficult if forged faces are mixed with real faces. To investigate these hypothesises, our user study focused on both two cases: cropped faces to eliminate surrounding contexts and full images with multi-face.}

\textbf{Evaluation of Image Realism.} We cropped the forged heads, which had been doubly extended from the faces, to ensure that the upper-half of each person was completely extracted. The participants were asked to view 200 forged head images and then provide feedback on each image's realism in the form of a score 1 to 5, corresponding to `clearly fake,' `weakly unreal,' `borderline,' `almost real,' and `clearly real.' As shown by the results in Fig.~\ref{fig:study_realistic}, the visual quality of the images in the OpenForensics dataset was highly evaluated by most of the participants. That is, the forged faces in the OpenForensics dataset were judged to be the most realistic. Our dataset achieved the highest mean opinion score (MOS) 4.0, much higher than that of the second-best dataset Celeb-DF (3.2). The DeeperForensics and DFDC datasets had medium-quality images (MOS of 2.8). The FaceForensics++ dataset had the most unrealistic images (MOS of only 1.3).

\begin{figure}[t!]
\vspace{-5mm}
	\centering
    \includegraphics[width=1\linewidth]{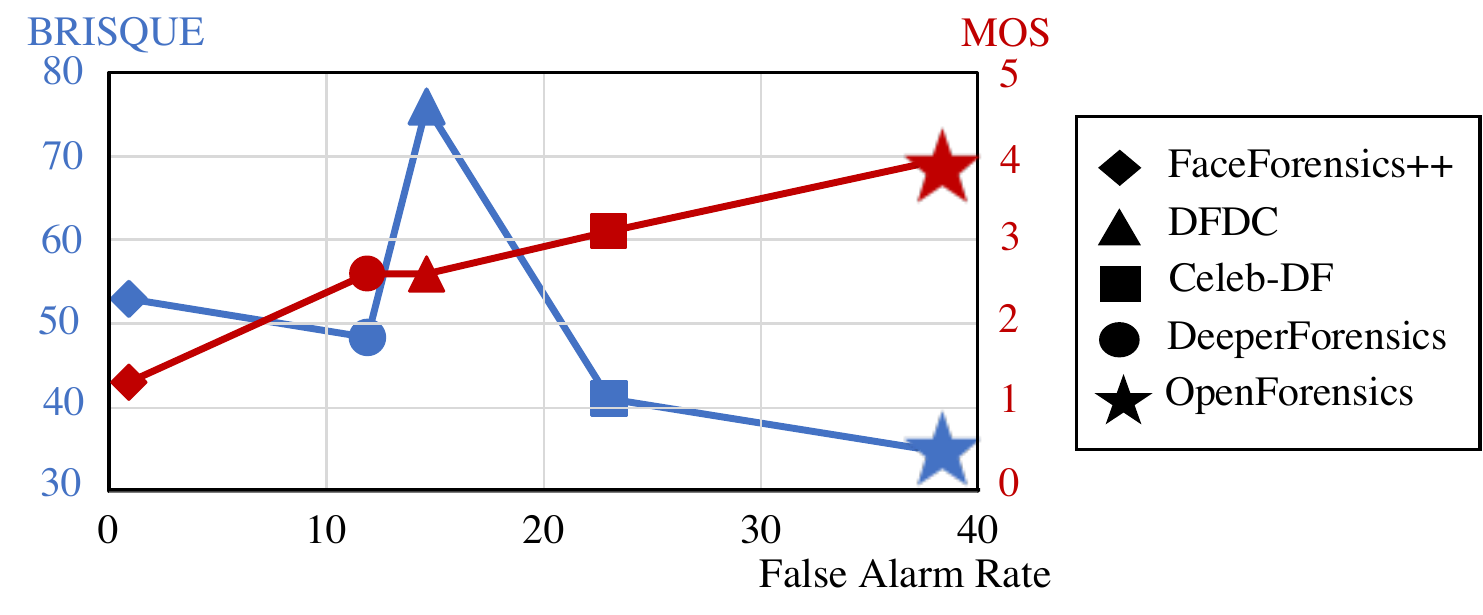}
	\caption{Correlation between visual properties and human ability to recognize forged faces. The ability to recognize forged faces depends on image realism (higher MOS is better) and visual quality (lower BRISQUE is better). False alarm rate is higher for images with higher quality and more realism, meaning that OpenForensics is the best dataset in terms of having realistic images.}
	\label{fig:subjective_test_corelation}
	\vspace{-2mm}
\end{figure}

\begin{figure}[t!]
	\centering
	\includegraphics[width=1\linewidth]{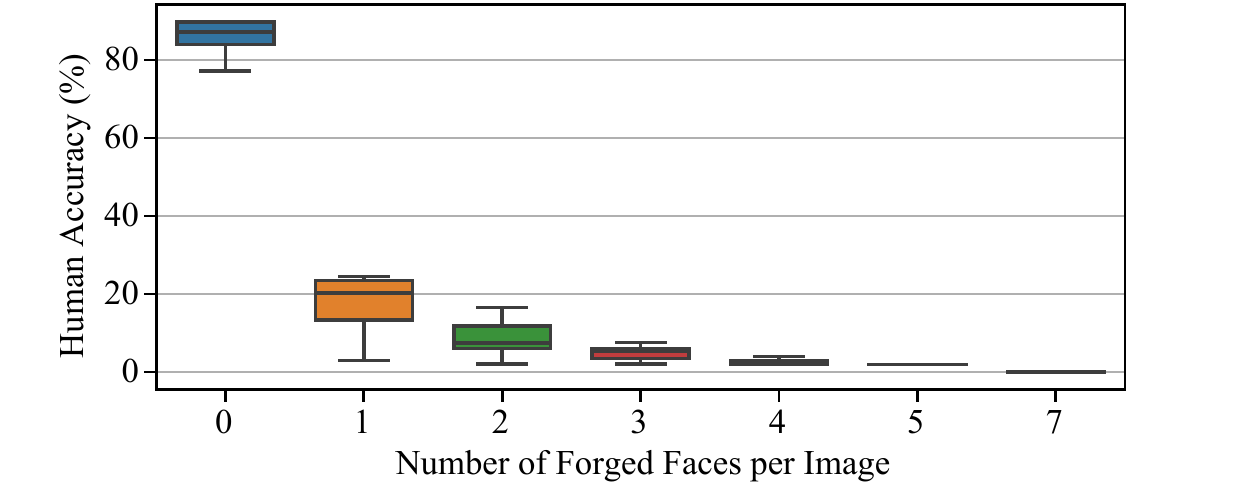}
	\caption{Human performance on multi-face forgery detection. Accuracy deceased as number of forged faces increased.}
	\label{fig:study_detection}
	\vspace{-3mm}
\end{figure}

\textbf{Human Performance on Face Forgery Classification.} We again cropped the heads similar to the cropping done for the evaluation of image realism. The participants were asked to view a mixture of 400 images randomly composed of pristine and forged heads with a ratio of 50:50. After viewing each image, the participants were asked whether the image was 'real' or 'fake.' As shown in Fig.~\ref{fig:subjective_test_detection_human_accuracy}, the participants had the most trouble distinguishing between the real and fake images in the OpenForensics dataset. This is evidenced by the OpenForensics dataset having the lowest overall accuracy (59.7\%), followed by Celeb-DF (68.7\%), DFDC (72.0\%), FaceForensics++ (82.0\%), and DeeperForensics (82.9\%,). The graph also shows that both experts and non-experts had difficulty distinguishing between the real and fake images in our dataset. \highlight{It is interesting that although experts could recognize fake faces better than non-experts, they incorrectly identified real faces with low quality, low resolution, or low contrast (\ie, FaceForensics++ dataset). We attribute this to their overconfidence and their belief that GANs might generate such faces, leading to misidentification.}

Figure \ref{fig:subjective_test_corelation} illustrates the correlation between the visual properties and the human ability to recognize forged faces. The ability to recognize forged faces depends on image realism, resulting in an increased false alarm rate as realism improves (\ie, as the MOS increases). The graph shows that a large number of participants misclassified forged faces in the OpenForensics dataset as real faces. The OpenForensics dataset had the highest MOS (4.0) and the highest false alarm rate (34.6\%). The figure also shows that the BRISQUE score~\cite{Mittal-ASILOMAR2011} of the OpenForensics dataset was the lowest (35.2), which indicates that the images in our dataset have the best visual quality. Reducing image quality (\ie, increasing the BRISQUE score) would affect human observation, resulting in a lower false alarm rate.

\textbf{Human Performance on Multi-Face Forgery Detection.} The participants were asked to view a set of 160 images, each with multiple persons and each consisting of both pristine and forged faces randomly selected, of only pristine faces, or of only forged faces. They were asked to identify the number of forged faces in each image. Figure \ref{fig:study_detection} shows that detection accuracy was the highest (86\%) when there were no forged faces in the images and tended to drop as the number of forged faces increased. \highlight{This can be explained that when there are many faces in an image, participants tend to less carefully check each face and guess that all the faces are real. That explains why the accuracy is high when all faces are real while it significantly reduces when forged faces exist.} Indeed, when the number exceeded 7, accuracy dropped to 0\%. Even people find it extremely difficult to identify forged faces among mixture of pristine and forged faces on in-the-wild images, highlighting the challenge of our OpenForensics dataset.


\section{Benchmark Suite}
\label{sec:benchmark}

\subsection{Baseline Methods}

We conducted a competitive benchmark for multi-face forgery detection and segmentation. To this end, we trained and evaluated the latest instance detection and segmentation methods in various scenarios. The methods were MaskRCNN~\cite{Kaiming-ICCV2017}, MSRCNN~\cite{Huang-CVPR2019}, RetinaMask~\cite{Fu-2019}, YOLACT~\cite{Bolya-ICCV2019}, YOLACT++~\cite{Bolya-PAMI2020}, CenterMask~\cite{Lee-CVPR2020}, BlendMask~\cite{Chen-CVPR2020}, PolarMask~\cite{Xie-CVPR2020}, MEInst~\cite{Zhang-CVPR2020}, CondInst~\cite{Tian-ECCV2020}, SOLO~\cite{Wang-ECCV2020}, and SOLO2~\cite{Wang-NeurIPS2020}. MaskRCNN and MSRCNN are well-known two-stage models that perform detect-then-segment slowly. The YOLACT ones~\cite{Bolya-ICCV2019, Bolya-PAMI2020} are early single-stage models aimed at real-time performance. The remaining methods are widely used modern single-stage models that overcome accuracy and processing time problems. Among them, the SOLO ones~\cite{Wang-ECCV2020, Wang-NeurIPS2020} directly output masks without computing bounding boxes.  


All the methods were used with the same backbone (FPN-ResNet50~\cite{Lin-CVPR2017, He-CVPR2016}) to make the comparison fair. We trained models on PCs with 32 GB of RAM and a Tesla P100 GPU. The models were initialized with ImageNet weights~\cite{Krizhevsky-NIPS2012} and trained on our training set for 12 epochs. The base learning rate was decreased by $1/10$ at the $8^{th}$ and $11^{th}$ epochs. Other settings were in accordance with the default public configurations provided by the authors.

\begin{table*}[t!]
\vspace{-5mm}
\caption{Benchmark results for multi-face forgery detection and segmentation on test-dev set. Higher AP is better while lower oLRP error is better. Best and second-best results are shown in \textcolor{blue}{blue} and \textcolor{red}{red}, respectively.}
\label{table:benchmark_testdev}
\resizebox{1\linewidth}{!}{
\begin{tabular}{lc|cccc|cccc|cccc|cccc}
\hline
\multicolumn{1}{c}{\multirow{2}{*}{\textbf{Method}}} & \multicolumn{1}{c}{\multirow{2}{*}{\textbf{Year}}} & \multicolumn{8}{|c|}{\textbf{Multi-Face Forgery Detection}} & \multicolumn{8}{|c}{\textbf{Multi-Face Forgery Segmentation}} \\ \cline{3-18}
\multicolumn{1}{c}{} & \multicolumn{1}{c}{} &
\multicolumn{1}{|c}{\textbf{AP$\uparrow$}} &
\multicolumn{1}{c}{\textbf{AP$_S$$\uparrow$}} &
\multicolumn{1}{c}{\textbf{AP$_M$$\uparrow$}} &
\multicolumn{1}{c}{\textbf{AP$_L$$\uparrow$}} &
\multicolumn{1}{|c}{\textbf{oLRP$\downarrow$}} & \multicolumn{1}{c}{\textbf{oLRP$_{Loc}$$\downarrow$}} & \multicolumn{1}{c}{\textbf{oLRP$_{FP}$$\downarrow$}} &
\multicolumn{1}{c}{\textbf{oLRP$_{FN}$$\downarrow$}} &
\multicolumn{1}{|c}{\textbf{AP$\uparrow$}} &
\multicolumn{1}{c}{\textbf{AP$_S$$\uparrow$}} &
\multicolumn{1}{c}{\textbf{AP$_M$$\uparrow$}} &
\multicolumn{1}{c}{\textbf{AP$_L$$\uparrow$}} &
\multicolumn{1}{|c}{\textbf{oLRP$\downarrow$}} & \multicolumn{1}{c}{\textbf{oLRP$_{Loc}$$\downarrow$}} & \multicolumn{1}{c}{\textbf{oLRP$_{FP}$$\downarrow$}} &
\multicolumn{1}{c}{\textbf{oLRP$_{FN}$$\downarrow$}} \\ \hline
MaskRCNN~\cite{Kaiming-ICCV2017} & ICCV 2017 & 79.2 & 29.9 & 80.2 & 79.5 & 24.3 & 9.5 & 2.7 & \textcolor{red}{4.0} & 83.6 & 16.1 & 82.1 & 85.8 & 21.2 & 7.6 & 3.0 & \textcolor{blue}{4.2}  \\
MSRCNN~\cite{Huang-CVPR2019} & CVPR 2019 & 79.0 & 29.5 & 80.1 & 79.5 & 24.3 & 9.6 & 2.7 & \textcolor{blue}{3.8} & 85.1 & 16.8 & 84.2 & 86.8 & 21.1 & 7.7 & 2.6 & \textcolor{red}{4.4} \\
\hline
RetinaMask~\cite{Fu-2019} & arXiv 2019 & 80.0 & 30.9 & 80.2 & 80.7 & 24.2 & 9.0 & 3.0 & 4.6 & 82.8 & 16.4 & 80.6 & 85.1 & 22.6 & 8.1 & 2.9 & 4.9 \\
YOLACT~\cite{Bolya-ICCV2019} & ICCV 2019 & 68.1 & 12.5 & 67.1 & 69.3 & 37.2 & 13.4 & 6.3 & 8.7 & 72.5 & 3.1 & 67.0 & 75.7 & 34.0 & 11.4 & 6.4 & 8.7 \\
YOLACT++~\cite{Bolya-PAMI2020} & TPAMI 2020 & 72.9 & 20.9 & 73.4 & 73.6 & 31.5 & 12.1 & 4.0 & 5.8 & 77.3 & 6.5 & 73.9 & 80.0 & 28.2 & 10.0 & 3.9 & 6.5 \\
CenterMask~\cite{Lee-CVPR2020} & CVPR 2020 & \textcolor{red}{85.5} & \textcolor{red}{32.0} & 85.2 & \textcolor{red}{86.2} & 21.1 & 6.8 & 3.3 & 5.9 & 87.2 & 16.5 & 85.0 & 89.4 & 21.4 & 6.1 & 3.2 & 7.8 \\
BlendMask~\cite{Chen-CVPR2020} & CVPR 2020 & \textcolor{blue}{87.0} & \textcolor{blue}{32.7} & \textcolor{blue}{86.3} & \textcolor{blue}{88.0} & \textcolor{blue}{19.5} & \textcolor{blue}{6.2} & \textcolor{red}{2.4} & 6.2 & \textcolor{blue}{89.2} & \textcolor{blue}{19.8} & \textcolor{blue}{87.3} & \textcolor{blue}{91.0} & \textcolor{blue}{18.3} & \textcolor{blue}{5.4} & 2.5 & 6.3 \\
PolarMask~\cite{Xie-CVPR2020} & CVPR 2020 & 85.0 & 27.4 & \textcolor{red}{85.4} & 85.7 & \textcolor{red}{20.7} & \textcolor{red}{6.6} & 2.5 & 6.6 & 85.0 & 15.3 & 83.3 & 87.0 & 21.3 & 6.9 & 2.5 & 6.6 \\
MEInst~\cite{Zhang-CVPR2020} & CVPR 2020 & 82.8 & 26.0 & 82.7 & 83.4 & 23.8 & 7.6 & 4.1 & 6.8 & 82.2 & 13.9 & 81.5 & 83.3 & 25.0 & 8.1 & 4.0 & 7.2 \\
CondInst~\cite{Tian-ECCV2020} & ECCV 2020 & 84.0 & 29.4 & 83.6 & 84.8 & 20.8 & 7.4 & \textcolor{blue}{2.3} & 5.2 & \textcolor{red}{87.7} & \textcolor{red}{18.1} & 85.1 & \textcolor{red}{89.8} & \textcolor{blue}{18.3} & \textcolor{red}{5.9} & \textcolor{red}{2.4} & 5.3 \\
SOLO~\cite{Wang-ECCV2020} & ECCV 2020 & - & - & - & - & - & - & - & - & 86.6 & 15.4 & \textcolor{red}{85.6} & 88.4 & \textcolor{red}{20.0} & 6.6 & \textcolor{blue}{2.1} & 6.0 \\
SOLO2~\cite{Wang-NeurIPS2020} & NeurIPS 2020 & - & - & - & - & - & - & - & - & 85.1 & 13.7 & 83.7 & 87.1 & 21.5 & 7.1 & 3.1 & 5.8 \\
\hline
\end{tabular}
}
\end{table*}

\subsection{Evaluation Metrics}

We evaluated the methods using standard COCO-style average precision (AP)~\cite{Lin-ECCV2014}. We report the results for mean AP and AP on different scales ($AP_{S}$, $AP_{M}$, $AP_{L}$, where S, M, and L represent small, medium, and large objects). We also evaluated the methods using the localization recall precision (LRP) error~\cite{Oksuz-ECCV2018}. We report the results for mean optimal LRP (oLRP) and its error components including localization (oLRP$_{Loc}$), the false positive rate (oLRP$_{FP}$), and the false negative rate (oLRP$_{FN}$). 

\begin{figure}[t]
    \centering
    \includegraphics[width=1\linewidth]{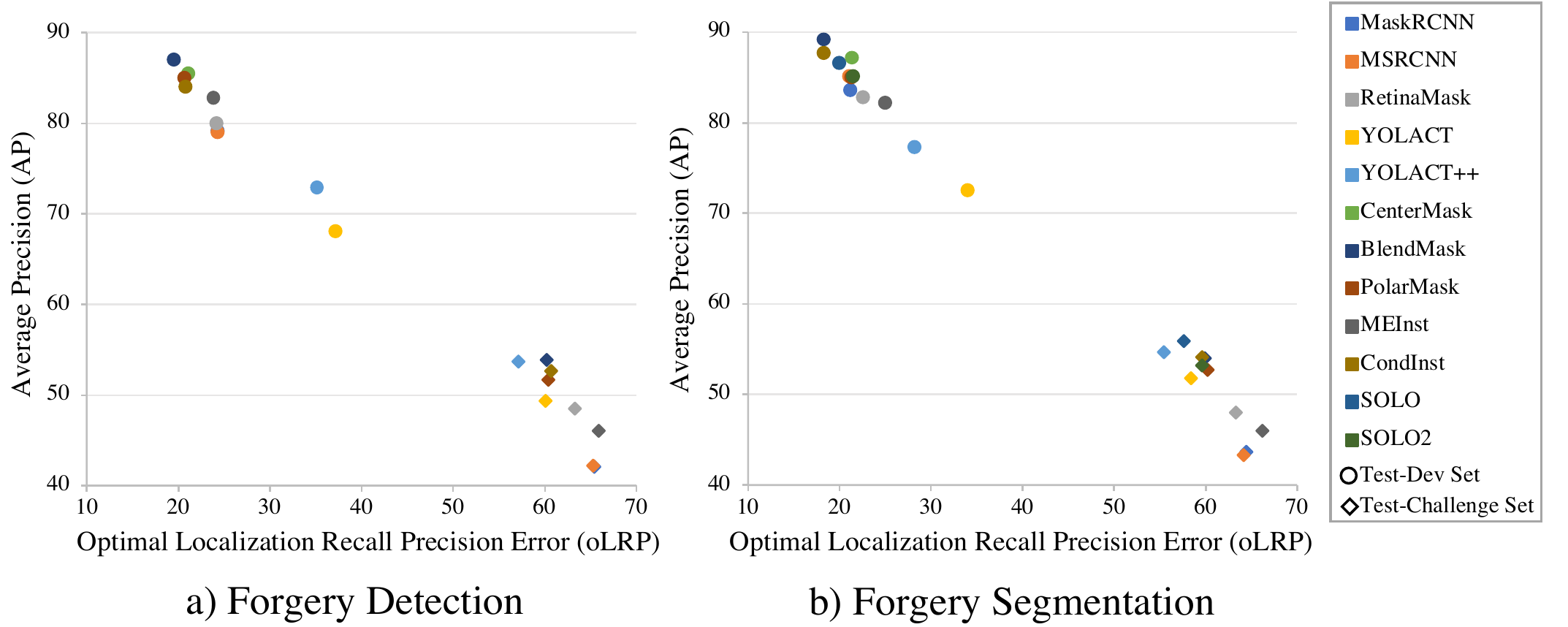}
    \caption{Benchmark results achieved by baseline methods for multi-face forgery multi-task on OpenForensics dataset (best viewed online in color with zoom-in). Test-dev set results reflect benchmark performance on standard images while test-challenge set results reflect robustness for unseen images. Lower oLRP error is better while higher AP is better. BlendMask had the best performance, and YOLACT++ was the most robust. Result for CenterMask on test-challenge set is out of the range and is shown in Table~\ref{table:benchmark_testchallenge}.}
    \label{fig:benchmark}
\end{figure}

\subsection{Overall Evaluation}

As shown in Fig.~\ref{fig:benchmark}, BlendMask had the best performance, with the highest AP and lowest oLRP error for both the detection and segmentation tasks on standard images. The other modern single-stage methods also had high performance, and the two-stage methods had medium performance. The YOLACT methods had the worst performance on both tasks because they are mainly focused on real-time processing. YOLACT++ and BlendMask were the most robust for unseen images.

\subsection{Multi-Face Forgery Detection Benchmark}

Table~\ref{table:benchmark_testdev} shows detailed results for the multi-face forgery detection task broken down by metric. They show that BlendMask had the best performance, achieving the highest AP (87.0) and the lowest oLRP error (19.5). BlendMask also achieved the highest AP for all object scales. The modern single-stage methods (\ie, BlendMask, PolarMask, and CondInst) had minor location errors and false positive rates while the two-stage methods (\ie, MaskRCNN and MSRCNN) had low false negative rates.


\begin{table*}[t!]
\vspace{-3mm}
\caption{Benchmark results for multi-face forgery detection and segmentation on test-challenge set. Higher AP is better while lower oLRP error is better. Best and second-best results are shown in \textcolor{blue}{blue} and \textcolor{red}{red}, respectively.}
\label{table:benchmark_testchallenge}
\resizebox{1\linewidth}{!}{
\begin{tabular}{lc|cccc|cccc|cccc|cccc}
\hline
\multicolumn{1}{c}{\multirow{2}{*}{\textbf{Method}}} & \multicolumn{1}{c}{\multirow{2}{*}{\textbf{Year}}} & \multicolumn{8}{|c|}{\textbf{Multi-Face Forgery Detection}} & \multicolumn{8}{|c}{\textbf{Multi-Face Forgery Segmentation}} \\ \cline{3-18}
\multicolumn{1}{c}{} & \multicolumn{1}{c}{} &
\multicolumn{1}{|c}{\textbf{AP$\uparrow$}} &
\multicolumn{1}{c}{\textbf{AP$_S$$\uparrow$}} &
\multicolumn{1}{c}{\textbf{AP$_M$$\uparrow$}} &
\multicolumn{1}{c}{\textbf{AP$_L$$\uparrow$}} &
\multicolumn{1}{|c}{\textbf{oLRP$\downarrow$}} & \multicolumn{1}{c}{\textbf{oLRP$_{Loc}$$\downarrow$}} & \multicolumn{1}{c}{\textbf{oLRP$_{FP}$$\downarrow$}} &
\multicolumn{1}{c}{\textbf{oLRP$_{FN}$$\downarrow$}} &
\multicolumn{1}{|c}{\textbf{AP$\uparrow$}} &
\multicolumn{1}{c}{\textbf{AP$_S$$\uparrow$}} &
\multicolumn{1}{c}{\textbf{AP$_M$$\uparrow$}} &
\multicolumn{1}{c}{\textbf{AP$_L$$\uparrow$}} &
\multicolumn{1}{|c}{\textbf{oLRP$\downarrow$}} & \multicolumn{1}{c}{\textbf{oLRP$_{Loc}$$\downarrow$}} & \multicolumn{1}{c}{\textbf{oLRP$_{FP}$$\downarrow$}} &
\multicolumn{1}{c}{\textbf{oLRP$_{FN}$$\downarrow$}} \\ \hline
MaskRCNN~\cite{Kaiming-ICCV2017} & ICCV 2017 & 42.1 & 11.8 & 46.2 & 40.5 & 65.4 & 13.6 & 29.3 & 40.0 & 43.7 & 4.7 & 44.3 & 44.0 & 64.4 & 11.8 & 29.4 & 41.2  \\
MSRCNN~\cite{Huang-CVPR2019} & CVPR 2019 & 42.2 & 11.8 & 45.9 & 40.8 & 65.3 & 13.7 & 29.6 & 39.9 & 43.3 & 5.2 & 44.6 & 43.5 & 64.1 & 11.8 & 30.4 & 39.6 \\
\hline
RetinaMask~\cite{Fu-2019} & arXiv 2019 & 48.5 & \textcolor{red}{12.8} & 51.0 & 48.1 & 63.3 & 12.6 & 33.2 & 34.6 & 48.0 & 4.7 & 46.5 & 49.7 & 63.3 & 11.8 & 30.9 & 38.0  \\
YOLACT~\cite{Bolya-ICCV2019} & ICCV 2019 & 49.4 & 5.6 & 49.6 & 50.3 & \textcolor{red}{60.1} & 15.3 & \textcolor{red}{23.2} & \textcolor{red}{29.9} & 51.8 & 1.4 & 47.2 & 54.6 & 58.4 & 13.5 & \textcolor{red}{23.4} & \textcolor{red}{30.1}  \\
YOLACT++~\cite{Bolya-PAMI2020} & TPAMI 2020 & \textcolor{red}{53.7} & 11.1 & 54.0 & \textcolor{blue}{54.8} & \textcolor{blue}{57.1} & 14.1 & \textcolor{blue}{19.7} & \textcolor{blue}{29.3} & \textcolor{red}{54.7} & 2.4 & 50.7 & \textcolor{blue}{57.9} & \textcolor{blue}{55.4} & 12.2 & \textcolor{blue}{20.0} & \textcolor{blue}{30.0}  \\
CenterMask~\cite{Lee-CVPR2020} & CVPR 2020 & 0.03 & 0.4 & 0.0 & 0.0 & 99.5 & 29.7 & 97.7 & 97.9 & 0.02 & 0.0 & 0.0 & 0.0 & 99.6 & 28.3 & 97.9 & 98.4  \\
BlendMask~\cite{Chen-CVPR2020} & CVPR 2020 & \textcolor{blue}{53.9} & \textcolor{blue}{13.5} & \textcolor{blue}{56.6} & \textcolor{red}{53.5} & 60.2 & \textcolor{blue}{10.6} & 26.5 & 37.4 & 54.0 & \textcolor{blue}{7.1} & \textcolor{red}{54.5} & 54.5 & 59.9 & \textcolor{blue}{9.8} & 26.4 & 38.4  \\
PolarMask~\cite{Xie-CVPR2020} & CVPR 2020 & 51.7 & 12.3 & 53.2 & 51.5 & 60.4 & \textcolor{red}{10.7} & 24.6 & 39.5 & 52.7 & 5.3 & 54.1 & 37.6 & 60.2 & 10.4 & 24.7 & 39.5  \\
MEInst~\cite{Zhang-CVPR2020} & CVPR 2020 & 46.1 & 8.6 & 49.9 & 44.9 & 65.9 & 12.4 & 34.6 & 39.7 & 46.0 & 3.8 & 49.0 & 45.2 & 66.2 & 12.6 & 34.8 & 39.8  \\
CondInst~\cite{Tian-ECCV2020} & ECCV 2020 & 52.7 & 12.6 & \textcolor{red}{55.3} & 51.8 & 60.7 & 11.5 & 28.3 & 35.3 & 54.1 & \textcolor{red}{6.5} & \textcolor{blue}{55.2} & 53.8 & 59.6 & \textcolor{red}{10.0} & 26.7 & 37.3  \\
SOLO~\cite{Wang-ECCV2020} & ECCV 2020 & - & - & - & - & - & - & - & - & \textcolor{blue}{55.9} & 3.9 & 53.3 & \textcolor{red}{57.3} & \textcolor{red}{57.6} & 11.3 & 24.6 & 33.0 \\
SOLO2~\cite{Wang-NeurIPS2020} & NeurIPS 2020 & - & - & - & - & - & - & - & - & 53.2 & 3.6 & 52.1 & 54.0 & 59.6 & 11.0 & 24.5 & 37.2 \\
 \hline
\end{tabular}
}
\vspace{-5mm}
\end{table*}

\subsection{Multi-Face Forgery Segmentation Benchmark}

With the emergence of explainable AI (XAI) technology~\cite{Doran-2017, Hagras-Computer2018, Rodney-MICCAI2020, ltnghia-IV2020}, it is useful to identify manipulated areas in detected faces. Therefore, we also evaluated segmentation performance. As shown in Table~\ref{table:benchmark_testdev}, for the multi-face forgery segmentation task, the trends in the ranking of method performance are similar to those for the detection task. BlendMask had the best segmentation performance, with AP of almost 90 and an oLRP error of approximately 18 for the test-dev set. 

Images in the real world obviously contain human faces of various sizes. It is thus essential to investigate detection and segmentation abilities on different scales. Table~\ref{table:benchmark_testdev} shows that all the baseline methods achieved high performance for only medium-size and large faces. Performance decreased with the face size, resulting in a marginal difference between small faces and medium/large faces in both detection and segmentation. These results illustrate the challenges of our OpenForensics dataset, which consists of enormous face sizes.

\highlight{Similar to the detection task, we found that single-stage methods, which are based on dense detection, have fewer FP errors while the two-stage ones, which are based on sparse detection, have fewer FN errors. Therefore, the development of post-processing using NMS and the improvement of RPN, respectively, can help to improve forgery detectors.}

\subsection{Robustness Evaluation}

We conducted experiments to evaluate the robustness of the methods on our test-challenge set, which simulates scenarios in the real world. Table~\ref{table:benchmark_testchallenge} shows that YOLACT++ and BlendMask were the most robust methods for unseen images. 
CenterMask was the least robust method, which is attributed to its results containing a lot of noise, resulting in extremely high false positive and false negative rates.

Tables~\ref{table:benchmark_testdev} and \ref{table:benchmark_testchallenge} show a substantial drop in performance for all methods for unseen images, which are beyond the distribution of the training set. Although existing methods can work well on standard images, their robustness is weak for unseen images. Even leading forgery-identification methods in the deep learning era remain limited and cannot yet effectively address real-world challenges (Top-1: $AP<60$ on test-challenge set). Hence, \textit{multi-face forgery detection and segmentation problems in-the-wild are still far from being solved, leaving much room for improvement}. These results also illustrate the challenges of our OpenForensics dataset.


\section{Conclusion and Outlook}
\label{sec:conclusion}

As part of our comprehensive study on multi-face forgery detection and segmentation in-the-wild, we created a large-scale dataset. In-depth analysis of our OpenForensics dataset demonstrated its diversity and complexity. We also conducted an extensive benchmark by evaluating state-of-the-art instance segmentation methods in various experimental settings. We expect that our OpenForensics dataset will boost research activities in deepfake prevention. We intend to continue enlarging this dataset to accompany future developments in deepfake technology. 

Thanks to the rich annotations in our OpenForensics dataset, there are a number of foreseeable research directions that will provide a solid basis for forgery and general face studies, including fundamental research (\eg, weak/semi-supervised/self-supervised detection/segmentation, universal network for multiple tasks) and specific research (\eg, anti-forgery robustness detection, forgery boundary detection, forgery ranking, face anonymization, face detection/segmentation, facial landmark prediction).



\textbf{Acknowledgements. } This work was partially supported by JSPS KAKENHI Grants JP16H06302, JP18H04120, JP21H04907, JP20K23355, and JP21K18023, and by JST CREST Grants JPMJCR18A6 and JPMJCR20D3, including the AIP challenge program, Japan.


{
\small
\bibliographystyle{ieee_fullname}
\bibliography{short_bibtex} 
}

\end{document}